# Localisation of Numerical Date Field in an Indian Handwritten Document

S Arunkumar[1], Pallab Kumar Sahu[2], Sudeep Gorai[2], Kalyan Ghosh[3]
[1]Dept of Information Technology
[2]Dept of Computer Science and Engineering
[3]Dept of Electronics and Communication Engineering
Institute of Engineering and Management
Kolkata, India

*Abstract*— This paper describes a method to localise all those areas which may constitute the date field in an Indian handwritten document. Spatial patterns of the date field are studied from various handwritten documents and an algorithm is developed through statistical analysis to identify those sets of connected components which may constitute the date. Common date patterns followed in India are considered to classify the date formats in different classes. Reported results demonstrate promising performance of the proposed approach.

## I. INTRODUCTION

Many institutions, business organisations etc. face the problem of processing handwritten document .No successful work regarding the decipherment of unconstrained cursive handwriting has been reported till date [1]. Nevertheless, when focused on certain restricted applications of handwritten text like revealing the location certain numerical data (phone number, pin code...), work becomes quite interesting. The deciphering of the location of the 'date' field in a handwritten document is one such interesting work which has been illustrated in this paper. This may find huge industrial importance as many handwritten documents are required to be sorted or categorized according to the dates mentioned on it. Our proposed algorithm is an advancement to make these industrial or organizational works automated. This will allow additional advantage to fax, photocopy and scanning machines, where sorting handwritten documents based on dates (mentioned in it) could appreciably be made automated.

Works regarding the recognition of a given date information has been reported by many [2][3][4],each establishing a unique technique of its own. These algorithms however assume that the given input is a date field (i.e. the pixel locations of the 'date' field is already considered to be known). The challenging task remaining, however, is the detection or identification of those pixels from handwritten documents which may constitute the date field. Our paper focuses only on this challenging issue, so that those pixels which are extracted could be fed into the above mentioned algorithms for recognition, thus making our work a pioneering one in the field of Document Image Analysis.

In India, the most commonly followed date patterns are DD-MM-YY, DD/MM/YY and DD.MM.YY. There are more date patterns like DD-MM-YYYY, DD/MM/YYYY, DD.MM.YYYY etc. but our paper focuses only on the above three patterns. It could be convincingly said that the proposed algorithm to locate the former patterns could also be used to locate the later ones with slight alterations.

In this paper we necessitate that the spatial orientation of the connected components in a numerical date field follows a specific structure and can be exploited for the localisation task. We thus target to find all classified date fields in each and every text line of the handwritten document.

## II. OVERVIEW OF THE PROPOSED ALGORITHM

The proposed algorithm comprises of a series of processes (depicted by a flowchart shown in Figure I) which includes Pre-processing, Scrutinization of Eight Consecutive Connected Components (ECCC) and Further Classification of DD-MM-YY and DD.MM.YY. Each of these processes is discussed in detail in the subsequent sections of the paper.

Since our study demanded us to have a well maintained database, a database was created (for both training and testing) by scanning numerous handwritten documents of various individuals. Each of these images (documents written on white paper) were scanned at 600 dpi and stored in JPEG format.

A section is also devoted to demonstrate the outcome of our experimentation. All the results obtained, having been enunciated to corroborate our study.

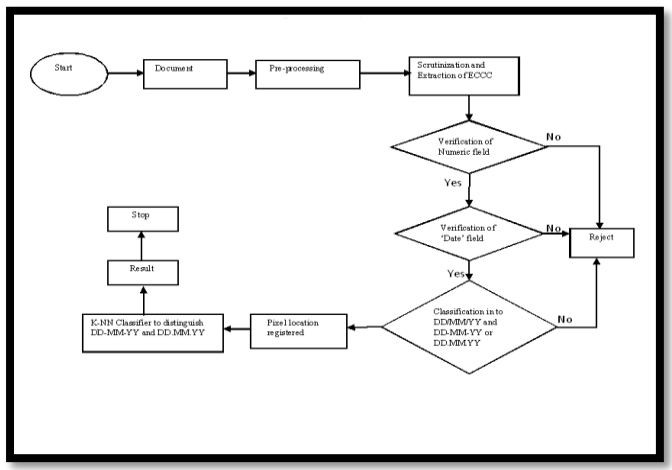

Figure I: the flowchart of the proposed algorithm.

### III. PREPROCESSING

Since our algorithm basically focuses on the scrutinization of the spatial arrangements of connected components and not on other aspects such as colour, texture etc, all the handwritten documents which are considered for statistical analysis or testing are converted to binary image such that the background is assigned a 'zero' pixel value and all the handwritten components are assigned a pixel value of 'one'. The overall image thus appears as shown in Figure III.

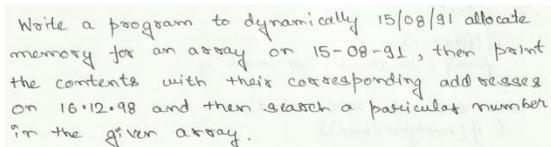

Figure II: showing the original document to be processed.

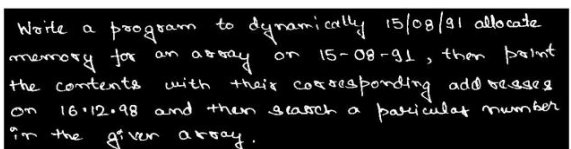

Figure III: showing the binary image of the converted document.

Once the document is converted into binary image (in the above mentioned way), all the text lines are extracted from it. Extraction of text lines implies grouping of connected components that belongs to the same line. For scrutinization of spatial features, the precise knowledge of these alignments is necessary. A histogram projection based text segmentation technique (inspired from [5]) is used.

### IV. SCRUTINIZATION OF EIGHT CONSECUTIVE CONNECTED COMPONENTS (ECCC)

The text lines extracted are then used for further examination. Since all the above specified classes (DD-MM-YY, DD/MM/YY and DD.MM.YY) deals with eight connected components so a group of eight consecutive connected components (ECCC) is extracted one at a time (say for example $C_1,C_2,C_3.....C_8$ ; where all $C_i$ belong to the same text line and $C_1$ is the first connected component of the ECCC). The widths of the minimum bounding rectangle enclosing these eight connected components are calculated and the maximum of these is found out and stored (say as $W_{max}$). A condition:- $X_{min}(C_{i+1}) > X_{min}(C_i)$ is used to eliminate instance(s) like the dot of 'i', noises, disoriented connected components (shown in Figure V and Figure VI.) etc. The goal now is to decipher whether the set of ECCC may constitute a date or not?

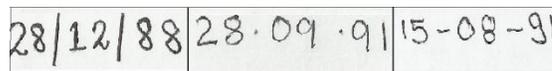

Figure IV: showing the three classes of date.

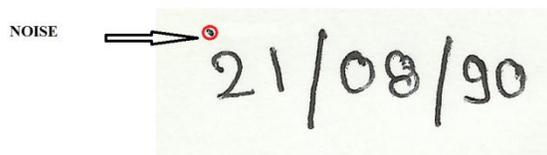

Figure V: showing the presence of noise (shown by an arrow mark).

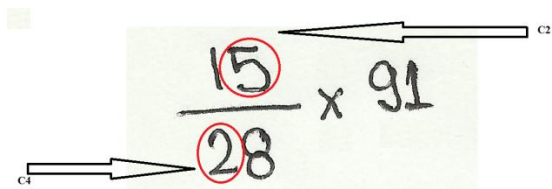

Figure VI: showing the case(s) eliminated when the condition $X_{min}(C_{i+1}) > X_{min}(C_i)$ is used; the connected component C2 and C3(denoted by arrow marks) violates the above condition, hence not detected as the desired ECCC.

The outline of the process is described as follows:-

*1) The horizontal interspatial distance between the above processed eight connected components is calculated*

(say for example $S_1,S_2,.....S_7$ ; where $S_i$ is the horizontal interspatial distance between $C_i$ and $C_{i+1}$ ). It is then checked to see that the value of no $S_i$ exceeds the value of 1.5times of $W_{max.}$ . This relation has been found out experimentally to avoid cases shown in Figure VII. It is a common observation that when dates are written, all the components representing it are within a certain horizontal interspatial distance from its neighbouring.

*2) If the set of eight consecutive connect components*

(say $C_i,C_{i+1},....C_{i+7}$) obeys with the conditions of the above step(Step I), then it is sent for further examination(Step III), else the next set (i.e. $C_{i+1}$ , $C_{i+2}$ ,....$C_{i+8}$) is considered and processed(Step I). This process goes on iteratively until all the set of eight consecutive is considered for a particular text line. When a text line is checked thoroughly (i.e. all the set of eight consecutive components is scrutinized), then the next text line is processed.

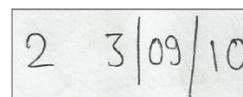

Figure VII: Incorrect formats of date:- a case that is avoided in our algorithm.

*3) Verification of numeric fields:-*

It could be easily learnt that in any classified format (as discussed above), the first, second, fourth, fifth, seventh and eighth constitute a numerical field. This process is inspired from [6] where features are defined to characterise the regularity of numerical fields. The feature vector is defined comprising of the following component f1, f2, f3, f4, f5, f6. Where for the set ECCC (say from $C_i$ to $C_{i+7}$) f1=$\frac{H(Ci+1)}{H(Ci)}$, f2=$\frac{Y(Ci+1)}{Y(Ci)}$, f3=$\frac{H(Ci+4)}{H(Ci+3)}$, f4=$\frac{Y(Ci+4)}{Y(Ci+3)}$, f5=$\frac{H(Ci+7)}{H(Ci+6)}$, f6=$\frac{Y(Ci+7)}{H(Ci+6)}$; where H represents height and Y represents Y co-ordinate of the centre of gravity of the minimum bounding rectangle enclosing the connected component. A training set of 250 documents is studied to learn the range values in which these features lie. These relations of the connected components with its immediate neighbours reveal features which may characterise it as a numerical field [6].

*4) Spatial Orientation of Numerical fields with respect to its Separators:-*

The above classified categories of date formats accommodate three types of separators, which are slash (/), dash (-) and dot (.). Learning of the spatial orientation of the numerical field with respect to its separators is the crux of our algorithm. A pattern is studied from a database of around 250 documents which thoroughly emphasizes on the localisation of the date field and classification of it into various categories of date format. Spatial features are extracted to classify the date format into DD/MM/YY, DD-MM-YY or DD.MM.YY and NON-DATE SET. Further classification is done to distinguish among DD-MM-YY and DD.MM.YY format.

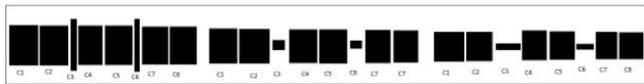

Figure VIII: showing a sample of the patterns of the minimum bounding rectangles of ECCC of all the three classes.

A feature vector is defined comprising of elements $Y_{min}(C_2)$, $Y_{min}(C_3)$, $Y_{min}(C_4)$, $Y_{min}(C_5)$, $Y_{min}(C_6)$, $Y_{min}(C_7)$, $Y_{max}(C_2)$, $Y_{max}(C_3)$, $Y_{max}(C_4)$, $Y_{max}(C_5)$, $Y_{max}(C_6)$, $Y_{max}(C_7)$; where $Y_{min}$ and $Y_{max}$ implies the minimum and maximum values of the Y co-ordinate of the minimum bounding rectangle.

Relationships are obtained among these features elements by training around 250 documents, these kinships are expressed (for the above defined classifications: DD/MM/YY, DD-MM-YY or DD.MM.YY and NON-DATE SET) in the form of mathematical inequalities (shown below).

*For Class DD/MM/YY:*
$Y_{min}(C_3) \leq Y_{min}(C_2) \leq Y_{max}(C_3)$
$Y_{min}(C_3) \leq Y_{max}(C_2) \leq Y_{max}(C_3)$
$Y_{min}(C_3) \leq Y_{min}(C_4) \leq Y_{max}(C_3)$
$Y_{min}(C_3) \leq Y_{max}(C_4) \leq Y_{max}(C_3)$
$Y_{min}(C_6) \leq Y_{min}(C_5) \leq Y_{max}(C_6)$
$Y_{min}(C_6) \leq Y_{max}(C_5) \leq Y_{max}(C_6)$
$Y_{min}(C_6) \leq Y_{min}(C_7) \leq Y_{max}(C_6)$
$Y_{min}(C_6) \leq Y_{max}(C_7) \leq Y_{max}(C_6)$

For Class DD-MM-YY or DD.MM.YY

$Y_{min}(C_2) \leq Y_{min}(C_3) \leq Y_{max}(C_2)$
$Y_{min}(C_2) \leq Y_{max}(C_3) \leq Y_{max}(C_2)$
$Y_{min}(C_4) \leq Y_{min}(C_3) \leq Y_{max}(C_4)$
$Y_{min}(C_4) \leq Y_{max}(C_3) \leq Y_{max}(C_4)$
$Y_{min}(C_5) \leq Y_{min}(C_6) \leq Y_{max}(C_5)$
$Y_{min}(C_5) \leq Y_{max}(C_6) \leq Y_{max}(C_5)$
$Y_{min}(C_7) \leq Y_{min}(C_6) \leq Y_{max}(C_7)$
$Y_{min}(C_7) \leq Y_{max}(C_6) \leq Y_{max}(C_7)$

The above eight cases of inequalities (defined for each of the above two categories i.e. DD/MM/YY and DD-MM-YY or DD.MM.YY) are used to categorise a set of ECCC into the above defined date formats. A set of ECCC falls into either of the categories if and only if it satisfies all the eight conditions defining that class. Those sets of ECCC which do not fall into either of the above categories are rejected and are labelled as 'NON- DATE' sets.

*5) Registering pixel locations:-*

Once the set of ECCC is labelled as 'date', the pixel location range (i.e. a rectangle having the co-ordinates $X_{min}, Y_{min}(C_i), X_{min}, Y_{max}(C_i), X_{max}, Y_{min}(C_{i+7}), X_{max}, Y_{max}(C_{i+7})$) is extracted. This region is now registered as 'date'. The output of a sample document (Figure IX) when processed is shown in Figure X.

The area localised is then sent for further classification if required (in case of DD-MM-YY and DD.MM.YY classes).

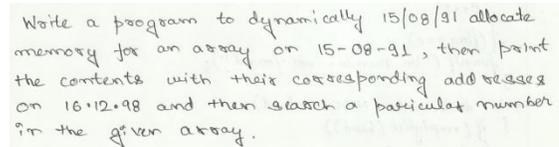

Figure IX: showing the image of a sample document that is used as an input for the above algorithm.

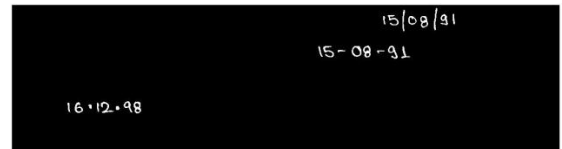

Figure X: showing the output image when the sample document (shown in Figure IX) is fetched as an input to our proposed algorithm (only the date fields are enunciated with pixel intensity value '1').

## V. FURTHER CLASSIFICATION OF DD-MM-YY AND DD.MM.YY FORMATS (OR CLASSES)

Both these classes of dates share common spatial attributes, hence categorising them based on the above features (or conditions) is not possible. The only distinguishing factor among them is the 3[rd] and 6[th] element of the set of ECCC.

A feature vector comprising of elements $W_{cc3}$ and $W_{cc6}$ is defined; $W_{cc3}$ and $W_{cc6}$ denote the width of the 3[rd] and 6[th] connected component respectively. A database comprising of 246 handwritten dates is trained to classify these classes based on the feature vector defined. Then KNN classifier (with value K=3) is used to classify the testing data (result shown in Table I).

Table I: Enunciating the results of the K-NN classifier used to distinguish between DD-MM-YY and DD.MM.YY format.

| No. of Documents | FAR (%) | FRR (%) | Efficiency (%) |
|---|---|---|---|
| 75 | 3.86 | 1.43 | 94.71 |
| 150 | 3.39 | 1.38 | 95.23 |
| 246 | 2.66 | 1.06 | 96.28 |

## VI. EXPERIMENT RESULTS

As mentioned earlier, the experiment was carried out (using Matlab 7.5.0.342, R2007b) on a database of 344 documents (157 of it were used for training and the remaining were used for testing). The results obtained are thus mentioned in a tabular form show in Table II.

Table II: Enunciating the results of date detection

| No. of Documents | FAR (%) | FRR (%) | Efficiency (%) |
|---|---|---|---|
| 50 | 12.00 | 6.00 | 82.00 |
| 100 | 10.00 | 4.00 | 86.00 |
| 187 | 9.09 | 3.20 | 87.71 |

## VII. CONCLUSION AND FUTURE WORKS

The proposed algorithm shows quite an interesting result. It can be clearly seen (from table I) that FRR (False Rejection Ratio) is far less than that of FAR (False Acceptance Ratio), moreover the percentage of efficiency increases as the number of documents considered(for testing) is increased. The high percentage of FAR is due to cases as depicted by Figure XI. FRR is basically due to illegible handwriting, deviations from the normal patterns (or syntax) and occurrence of double digits (Figure XII).

Since the localisation technique does not involve any recognition process, so the overall algorithm could be rated as quite simple and fast. As mentioned earlier this prescribed algorithm could be modified to localise more classes of dates.

Future works include studying similar patterns among alpha-numeric date formats and addressing the failure in localising dates (numerical) pregnant with 'double digits'.

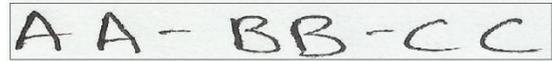

Figure XI: showing cases due to which FAR increases. The above script bears the same pattern as that of a date.

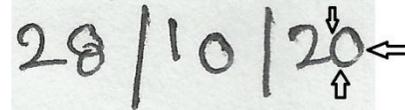

Figure XII: showing the case of double digits; the digits '2' and '0' are interconnected.